\documentclass[final,3p,times,twocolumn]{elsarticle}

\usepackage{blindtext, graphicx, amsmath, algorithm, algpseudocode, pifont, algcompatible, comment, layout, amsthm, amssymb}
\usepackage{enumitem}   
\usepackage{eso-pic}
\usepackage{booktabs}
\usepackage{float}
\usepackage{subfigure}

\usepackage{amssymb}
\usepackage{tabularx}

\usepackage[utf8]{inputenc}
\usepackage[english]{babel}
\usepackage{hyperref}

\usepackage{caption}
\usepackage{graphicx}
\usepackage{caption}

\journal{ICT Express}

\usepackage{listings}
\usepackage{xcolor}
\usepackage{tikz}
\definecolor{codegreen}{rgb}{0,0.6,0}
\definecolor{codegray}{rgb}{0.5,0.5,0.5}
\definecolor{codepurple}{rgb}{0.58,0,0.82}
\definecolor{backcolour}{rgb}{0.99, 0.99, 0.99}

\lstdefinestyle{mystyle}{
    backgroundcolor=\color{backcolour},   
    commentstyle=\color{codegreen},
    keywordstyle=\color{magenta},
    numberstyle=\tiny\color{codegray},
    stringstyle=\color{codepurple},
    basicstyle=\fontsize{6}{9}\selectfont\ttfamily,
    breakatwhitespace=false,         
    breaklines=true,                 
    captionpos=b,                  
    keepspaces=true,                 
    numbers=left,                    
    numbersep=5pt,                  
    showspaces=false,                
    showstringspaces=false,
    showtabs=false,                  
    tabsize=1,
    language=Python,
    morekeywords={as},  
}

\newcommand{\rev}[1]{{\textcolor{black}{#1}}}
\newcommand{\revv}[1]{{\textcolor{black}{#1}}}

\begin{document}

\begin{frontmatter}

\title{\rev{Towards} Storage-Aware Learning with Compressed Data\\ An Empirical Exploratory Study on JPEG}

\author{Kichang Lee$^{1,2}$}
\ead{kichang.lee@yonsei.ac.kr}
\author{JeongGil Ko$^{1}$}
\ead{jeonggil.ko@yonsei.ac.kr}
\author{Songkuk Kim$^{1}$\corref{cor}}
\ead{songkuk@yonsei.ac.kr}
\author{JaeYeon Park$^{3}$\corref{cor}}
\ead{jaeyeon.park@dankook.ac.kr}
\address{$^1$ School of Integrated Technology, Yonsei University, Seoul, Korea \\ $^2$ BK21 Graduate Program in Intelligent Semiconductor Technology, Yonsei University, Seoul, Korea \\ $^3$ Department of Mobile Systems Engineering, Dankook University, Yongin, Korea}
\cortext[cor]{Co-corresponding authors}
\begin{abstract}
\rev{On-device machine learning is fundamentally constrained by limited storage, especially in continuous data collection scenarios where sensor or vision streams accumulate rapidly. This paper empirically investigates storage-aware learning, characterizing the trade-off between data quantity and data quality under lossy compression. Using the CIFAR-10 dataset as a controlled benchmark, we systematically vary both the amount and the fidelity of training data to understand their joint impact on model performance. Our results reveal that (1) neither maximizing quantity nor quality alone yields optimal accuracy, emphasizing that the optimal trade-off between them depends nonlinearly on the available storage budget, and (2) data samples exhibit differential sensitivity to compression, motivating a sample-wise adaptive compression policy. These findings challenge uniform data-retention strategies such as naive data dropping or fixed-rate compression, and establish a foundation for adaptive, storage-efficient learning systems on resource-limited devices. This work opens new directions toward generalizable, storage-aware on-device intelligence.}

\end{abstract}
\end{frontmatter}

\pagestyle{plain}


\section{Introduction}
\vspace{-1.25ex}
\label{sec:intro}
The rapid advancement of technology over the past few decades has fueled the development of a wide range of applications powered by machine learning. Building on this momentum, the evolution of mobile and embedded devices has ushered in a new era where intelligent systems can operate efficiently under stringent resource constraints~\cite{gim2022memory}. This shift has brought machine intelligence closer to the data source, unlocking a new paradigm that moves beyond mere model execution: \textit{on-device model training and personalization}~\cite{park2023attfl}.

This paradigm addresses the inherent heterogeneity of real-world data from diverse contexts, environments, and hardware, which often undermines the performance of generic models. By continuously learning from individual user data, on-device adaptation mitigates these challenges\revv{~\cite{shin2024effective, park2024fedhm}}, improving accuracy, robustness, and responsiveness for a more personalized user experience.


\rev{However, effective on-device training faces practical hurdles beyond compute. Prior work largely targets CPU, memory, and energy constraints~\cite{gim2022memory, shin2024effective, zhu2024device, lee2024gmt}, but storage pressure is an equally critical and underexplored bottleneck\revv{~\cite{vopson2020information,ergen2024edge}}. Even as mobile storage grows, continuous sensing (cameras, microphones, multi-channel sensors) can quickly exhaust capacity\revv{~\cite{ouyang2024admarker,wang2025systematic,zhao2025lightweight}}. For example, 44.1 kHz 24-bit stereo audio requires \(\approx 22\)~GB/day (\(\sim\!660\)~GB/month); combined with high-resolution image streams or multi-sensor data, the demand escalates rapidly, making storage a first-order limitation for mobile and embedded learning.}

\rev{Such storage constraints motivate the development of strategies for managing excessive data accumulation. In previous works, typically two complementary strategies are involved}: (1) controlling the \textit{quantity} of stored data and (2) improving its \textit{quality}. Quantity-focused approaches prune samples to reduce dataset size, with early methods relying on static heuristics to discard less informative data~\cite{shin2022fedbalancer}. However, the limited adaptability of such heuristics has led to more dynamic techniques, including uncertainty-based and representative sampling~\cite{sener2017active}. Quality-focused approaches, on the other hand, aim to compress individual samples, often using lossy schemes that allow controlled information loss for substantial storage savings~\cite{wang2022memory}. In practice, these two strategies are interdependent, and addressing them in isolation is rarely sufficient for optimal performance.


Given these two orthogonal approaches, in practical applications, storage constraints are not governed by a single factor but rather by a complex interplay between data quantity and quality. For any given storage budget, a fundamental trade-off exists: one can either store a large volume of low-quality (highly compressed) data or a smaller set of high-quality (less compressed) data. 
\rev{However, despite the growing prevalence of storage-limited on-device learning, existing studies have rarely examined how this trade-off concretely influences model performance. Prior efforts have treated sample reduction (quantity)~\cite{sener2017active,paul2021deep} and compression (quality)~\cite{ghazvinian2019impact, mandelli2020training} as independent problems, overlooking their coupled effects on learning dynamics. Motivated by this overlooked yet practically decisive challenge, this work takes a first step toward systematically characterizing how storage constraints shape learning behavior. Specifically, our objective is to investigate (1) the impact of storage limitations on model training and (2) the necessity of balancing data quality and quantity to optimize learning outcomes under such constraints.}


To effectively explore this quantity vs. quality trade-off, it is crucial to recognize that while the benefits of increasing data quantity are well-established in machine learning, the consequences of altering data quality are far more nuanced and underexplored. Thus, a critical first step towards enabling a strategic balance is to characterize how quality degradation (via compression) fundamentally impacts the learning process, and critically, whether this impact is uniform across all data. Accordingly, this paper addresses the following foundational research questions \rev{(\textbf{RQ}s)}:

\vspace{0.1ex}
\rev{\noindent{$\bullet$} \textbf{RQ1 (Training under a storage budget):} Given a fixed storage budget, is there a definitive priority between data quality and quantity, or is a strategic balance needed to optimize generalization performance?}


\vspace{0.1ex}
\noindent{$\bullet$} \textbf{RQ2 (Impact of compression)}: What is the precise impact of data compression and the associated degradation in quality on the behavior of deep learning models?

\vspace{0.1ex}
\noindent{$\bullet$} \textbf{RQ3 (Differential Sensitivity)}: Are all data samples equally sensitive to compression, or do they exhibit varying levels of sensitivity?

\vspace{0.1ex}
As an exploratory study, this work investigates these questions through a series of targeted experiments on a standard computer vision dataset. We note that the goal of this work is to identify a novel problem domain and propose a feasible direction for future research in storage-aware learning. Specifically, this paper makes the following contributions:
\begin{itemize}[leftmargin=*]
    \item We highlight the necessity of studying model training under practical, storage-limited scenarios, an often-overlooked constraint in on-device AI.
    \item We empirically demonstrate that naive strategies, such as uniform data drop or applying a uniform compression rate, are suboptimal for maximizing performance under a fixed storage budget.
    \item We reveal individual data samples exhibit differential sensitivity to compression, which empirically shows the feasibility and potential of developing a sample-wise adaptive compression strategy.
\end{itemize}

\rev{The remainder of this work is structured as follows. Section~\ref{sec:relwork} reviews related work and Section~\ref{sec:prelim} presents the empirical analysis of the quantity-quality trade-off, the impact of compression, and per-sample sensitivity. We discusses implications, limitations, and future work in Section~\ref{sec:discussion} and conclude the work in Section~\ref{sec:conclusion}.}

\section{Related Work}
\vspace{-1.25ex}
\label{sec:relwork}
\rev{Scaling laws indicate that increasing data, model size, and compute improves generalization~\cite{kaplan2020scaling}. This creates a dilemma for on-device learning: prior work largely tackles compute limits (CPU, RAM), yet continuous collection makes storage a first-class constraint as raw streams can overwhelm device capacity~\cite{ouyang2024admarker}.}
\rev{Traditional ``store-everything'' pipelines are thus impractical, motivating data retention strategies that selectively preserve informative content. Two main levers are (1) sample selection to control \emph{quantity} and (2) lossy compression to control \emph{quality}. Compression reduces storage at the cost of fidelity~\cite{wang2022memory}, while dropping removes samples~\cite{paul2021deep}, yielding a non-trivial trade-off among compression artifacts, data elimination, and final accuracy.}

\noindent\textbf{\rev{Research Gap and Positioning.}}
\rev{Most prior work focus on \emph{quantity}-focused selection (e.g., active/core-set, early-example retention) or \emph{quality}-focused training with compression in isolation, offering limited insight into their \emph{joint effect} under a fixed storage budget. Moreover, \emph{per-sample} sensitivity to compression (central to adaptive policies) has not been systematically quantified. This paper fills these gaps by empirically characterizing the budgeted quantity-quality trade-off, showing the suboptimality of uniform policies, and revealing sample-wise compression sensitivity that motivates adaptive, storage-aware training.}

\section{Empirical Analysis}
\vspace{-1.25ex}
\label{sec:prelim}
This section presents an empirical analysis designed to address the foundational research questions outlined in Section~\ref{sec:intro}. Through a series of targeted experiments, we investigate the trade-off between data quality and quantity, the impact of data compression on model behavior, and the differential sensitivity of individual samples to quality degradation. To support our analysis, we conduct experiments using a 3-layer convolutional neural network (CNN) with three convolution layers, trained on the CIFAR-10 dataset for 30 epochs. We deliberately chose this standard dataset and simple architecture to create a controlled environment where the effects of data quality and quantity could be clearly isolated; thus, minimizing confounding factors from more complex models or data distributions. 
\vspace{-1.25ex}

\subsection{Training under a storage budget (RQ1)}
\begin{figure}[t!]
    \centering    
    \includegraphics[width=.95\linewidth]{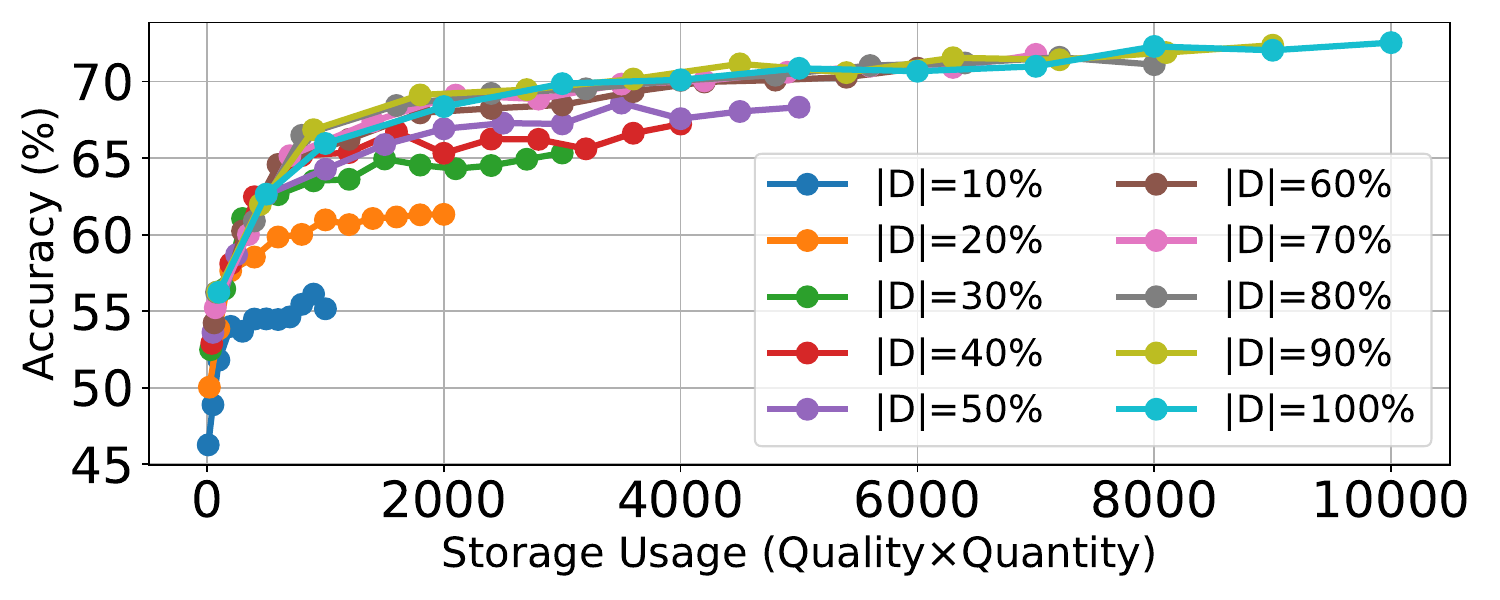}
    \vspace{-2ex}
    \caption{Model accuracy as a function of storage usage under various data quantity and quality settings. Each line represents a fixed data quantity (10\% to 100\%), where points along a line indicate increasing data quality with higher storage usage.}
    \label{fig:prelim1}
\end{figure}

\begin{figure}[t!]
    \vspace{-2ex}
    \centering    
    \includegraphics[width=.95\linewidth]{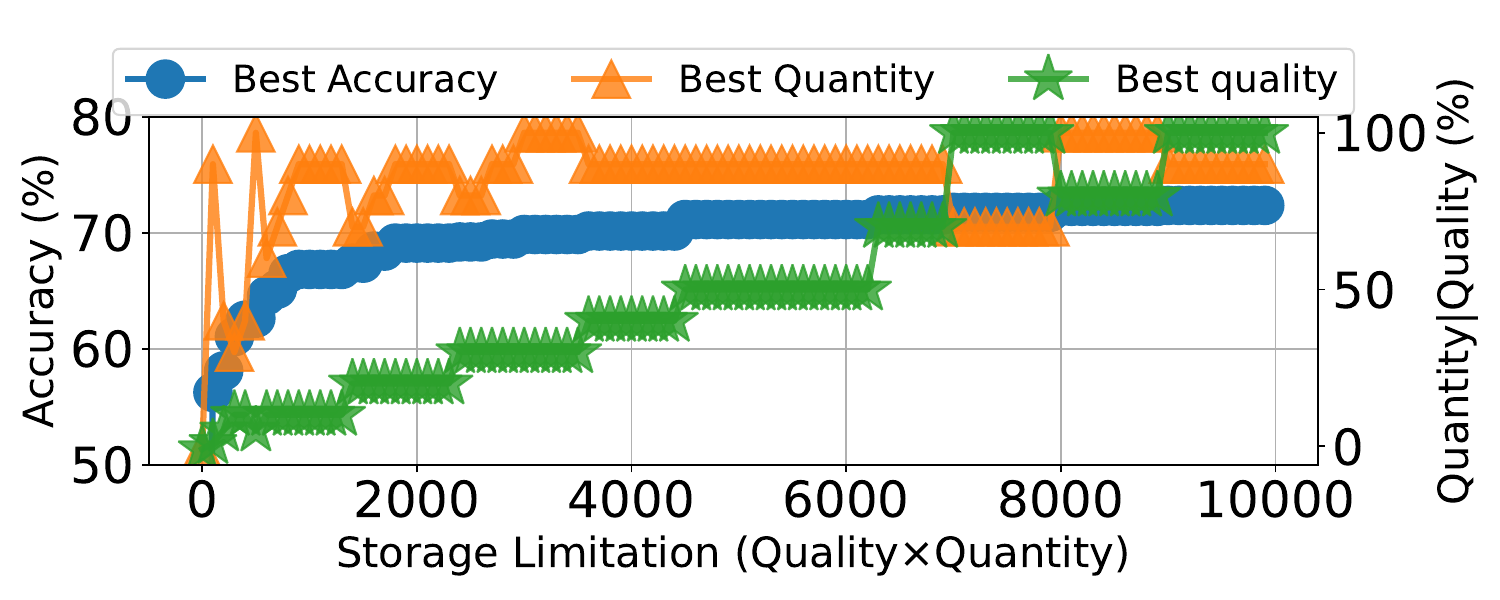}
    \vspace{-2ex}
    \caption{Best model accuracy and the corresponding data configuration for various storage limitations. For each storage budget, the plot shows the highest achievable accuracy ($\textcolor{blue}{\bullet}$), along with the specific data quantity ($\textcolor{orange}{\blacktriangle}$) and quality ($\textcolor{green}{\star}$) settings required to obtain that performance.}
    \vspace{-2ex}
    \label{fig:prelim2}
\end{figure}


We begin our investigation by examining the trade-off between data quantity ($|D|$) and data quality ($Q$) as the JPEG quality factor ($\%$) in relation to model performance. To this end, we train a CNN using varying amounts of training data (i.e., quantity) and different levels of JPEG compression (i.e., quality). We exploit JPEG\rev{,} given its high compression ratio, a feature of this standard lossy method, is essential for our study on storage-constrained learning. For this experiment, we apply a fixed compression rate uniformly across all training samples. In our setup, higher compression corresponds to lower-quality data, as lossy compression discards subtle or infrequent information in the input. We define an approximation of storage usage as the product of quantity and quality, $|D|$$\times$$Q$$\in$$(0, 10000]$.

Figure~\ref{fig:prelim1} presents the model's accuracy on the original (uncompressed, $Q$=$|D|$=100\%) test set as a function of storage usage, across different data quantities. As expected, model accuracy improves as storage usage increases, reflecting the benefit of higher quality or larger training datasets. However, a key observation is that under a fixed storage budget, the optimal quantity-quality combination varies; suggesting that maximizing either one alone does not assure the best performance. \rev{Empirically, under tight budgets ($<2000$), increasing \emph{quantity} tends to outperform prioritizing \emph{quality}, i.e., more (lower-quality) samples are typically optimal.}

Figure~\ref{fig:prelim2} further illustrates this by plotting the highest achieved accuracy and the corresponding optimal configuration (i.e., the best quantity–quality pair) for a range of storage constraints. As shown, the ideal trade-off between quantity and quality depends heavily on the available storage, which may be restricted by hardware limitations, system-level policies, or application-specific constraints. These findings emphasize the importance of \textit{\textbf{jointly optimizing both dimensions when designing storage-constrained learning systems}}.

\subsection{Impact of Data Compression (RQ2)}
\label{subsec:prelim2}
\begin{figure}[t!]
    \centering    
    \includegraphics[width=.85\linewidth]{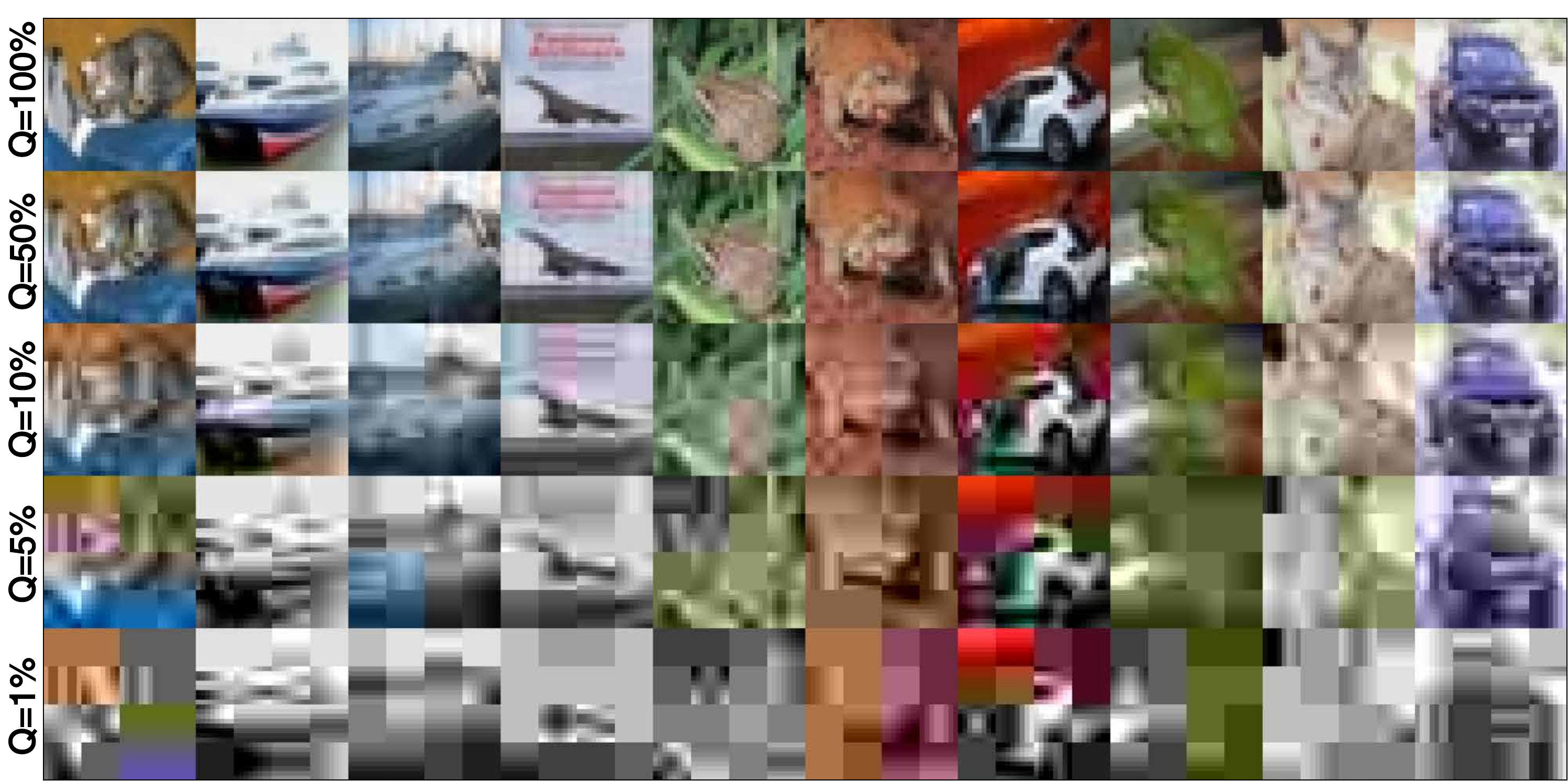}
    \vspace{-2ex}
    \caption{10 samples from the CIFAR10 dataset with different data quality (i.e., compression rate, $Q$).}
    \vspace{-2ex}
    \label{fig:samples}
\end{figure}
\rev{Next, to assess the impact of lossy compression on training data, we trained three CNNs on datasets at 5\%, 50\%, and 100\% quality and evaluated them on test sets compressed to multiple levels. Before reporting results, we present representative samples: Figure~\ref{fig:samples} shows the first 10 CIFAR-10 test images across qualities. As expected, image details degrade as compression increases; at extreme settings (e.g., \(Q{=}1,\,5\%\)), critical semantic information is heavily distorted or lost.}

\begin{figure}[t!]
    \centering    
    \includegraphics[width=.9\linewidth]{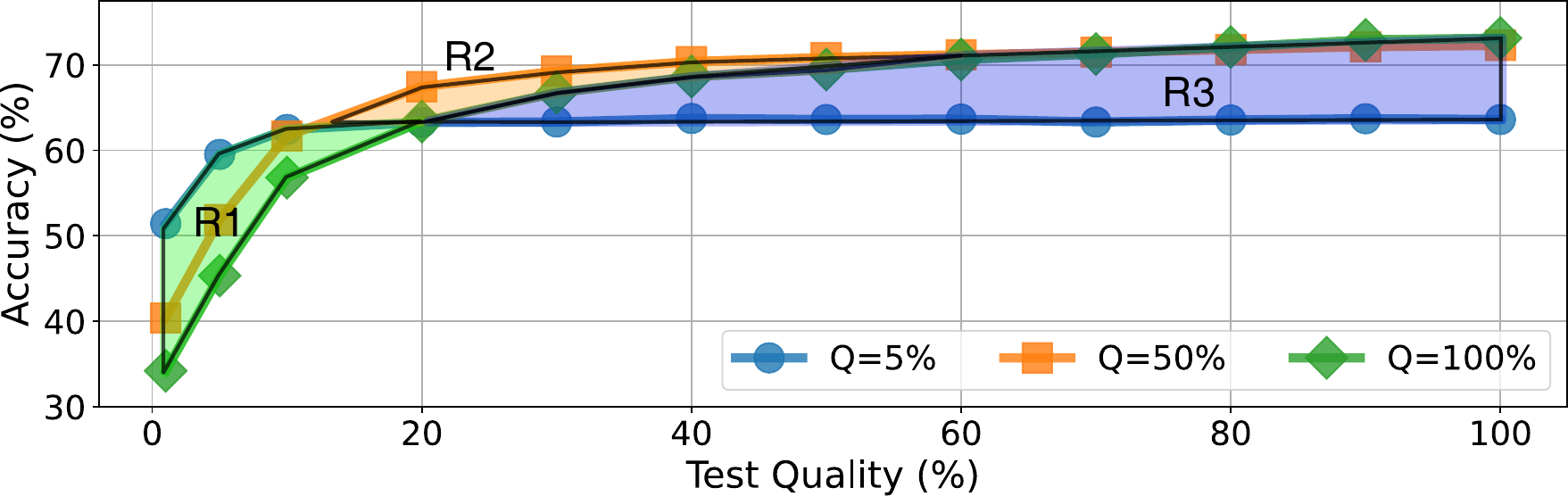}
    \vspace{-2ex}
    \caption{\rev{Test accuracy as a function of test data quality for models trained with different training data qualities ($Q$=5\%~(\textcolor{blue}{$\bullet$}), $Q$=50\%~(\textcolor{orange}{$\blacksquare$}), and $Q$=100\%~(\textcolor{green}{$\blacklozenge$})). The shaded regions, R1 and R2, indicate where models trained on 5\% and 50\% quality data, respectively, achieve optimal performance,  whereas region R3 denotes where higher-quality models ($Q$=50\%, $Q$=100\%) outperform the others.}}
    \vspace{-2ex}
    \label{fig:prelim3}
\end{figure}

\rev{Figure~\ref{fig:prelim3} reports test accuracy across test-sample qualities for models trained at different compression levels. Let \(f_{Q}\) denote the model trained with quality \(Q\)\% data and \(Acc_{Q}\) its accuracy. Three regions emerge (shaded in the figure): R1 (low test quality), where \(f_{5}\) outperforms \(f_{100}\) \((Acc_{5} > Acc_{100})\); R2 (mid quality), where \(f_{50}\) achieves the highest accuracy \((Acc_{50} > Acc_{100}, Acc_{5})\); and R3 (high quality), where \(f_{5}\) fails to improve further and higher-quality models \(f_{50}\) and \(f_{100}\) improve and lead \((Acc_{50}, Acc_{100} > Acc_{5})\).}

\begin{figure}[t!]
    \centering    
    \includegraphics[width=.9\linewidth]{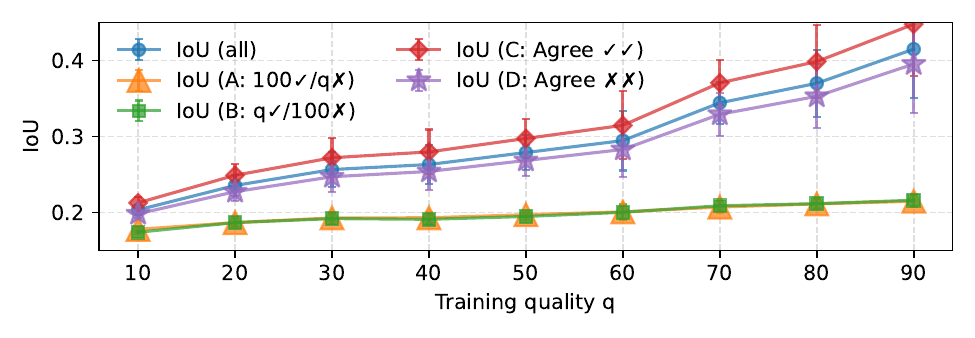}
    \vspace{-2ex}
    \caption{\rev{IoU between top-5\% SHAP masks at training quality $q$ and the same-seed baseline ($q{=}100$), averaged over five seeds. Lines summarize all samples and four cohorts: A ($100\checkmark/q\times$), B ($q\checkmark/100\times$), C (agree--correct), and D (agree--wrong). Higher IoU indicates more consistent explanations.}}
    \vspace{-3ex}
    \label{fig:iou}
\end{figure}

\rev{Since $f_{5}$ was unable to capture the rich features present in high-quality data during training, it performs poorly in R3. Notably, $f_{5}$ still achieves relatively high accuracy (above $60\%$) in R3, despite substantial loss of semantic information at very low quality levels (as exemplified in Fig.~\ref{fig:samples}). This stems from low-quality features being inherently embedded within high-quality data, allowing compressed models like $f_{5}$ to exploit minimal yet informative cues. Conversely, in R1 and R2, the model trained with original-quality data ($f_{100}$) performed worse than others. Consistent with R3, while $f_{100}$ can learn low-quality features, it primarily focuses on features unique to high-quality data. As a result, lower-quality training, specifically $f_{5}$ and $f_{50}$, yielded more robust generalization in R1 and R2, respectively.}

\rev{Figure~\ref{fig:iou} compares top-5\% SHAP masks from models trained at quality $q$ to baselines at $q{=}100$, averaged over five seeds and split by agreement cohorts. IoU decreases as $q$ drops, indicating a shift in relied-on evidence. This explains Figure~\ref{fig:prelim3}: in R3 high-quality tests, $f_{5}$ diverges from $f_{100}$ and underperforms; in R1 low-quality tests, $f_{5}$ aligns with compressed cues and fares better, while $f_{100}$ relies on fine details that vanish. Agreement cohorts show higher IoU, whereas disagreement cohorts A ($100\checkmark/q\times$) and B ($q\checkmark/100\times$) show low IoU, indicating different focal regions. These results support that stronger training-time compression changes feature reliance, preventing low-$q$ models from capturing the high-quality features used by $f_{100}$; conversely, $f_{100}$ underutilizes the low-frequency, artifact-tolerant cues emphasized by low-$q$ models, which explains its weaker performance on heavily compressed inputs.}

Based on these observations, we summarize three key findings: (1) Features extracted from low-quality data remain informative and useful (as seen in R3); (2) Relying solely on features exclusive to high-quality data can adversely affect robustness and generalization during inference (evident in R1 and R2); and (3) utilizing intermediately compressed data can offer performance advantages (demonstrated in R1 and R2). Motivated by these empirical results, we hypothesize that \textbf{\textit{it is possible to achieve an optimal balance between storage efficiency and model performance by adaptively selecting appropriate compression qualities}}.

\subsection{Differential Sensitivity to Compression (RQ3)}
\begin{figure}[t!]
    \centering
    \subfigure[Loss]{
    \includegraphics[width=0.47\linewidth]{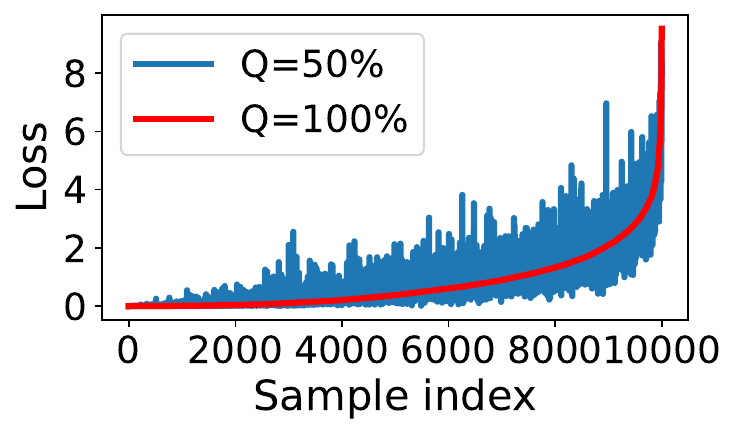}
    }
    \subfigure[$\Delta$Loss]{
    \includegraphics[width=0.47\linewidth]{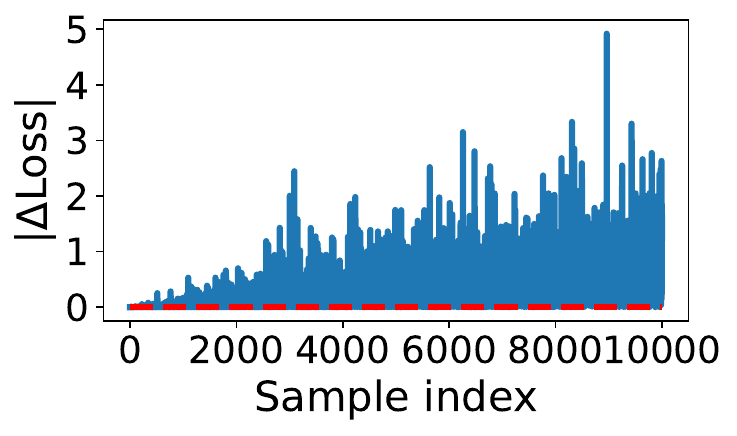}
    }
    \vspace{-3ex}
    \caption{Test loss comparison for original ($Q$=100) and compressed ($Q$=50) data, sorted by the original loss. (a) Per-sample loss values. (b) Magnitude of the loss difference between $Q$=100\% and $Q$=50\%}
    \vspace{-2ex}
    \label{fig:prelim4}
\end{figure}
\rev{Rephrasing the first finding in Section~\ref{subsec:prelim2}, some samples are accurately classified without high-quality-specific features (R1), while others are not (R3). This prompts the question: \textit{``Are all samples equally robust or sensitive to compression?''} To investigate, we compared each CIFAR-10 test sample’s loss at original quality with its loss under 50\% compression.}

\rev{Figure~\ref{fig:prelim4}~(a,b) plot the test losses and their differences ($|l_{100}-l_{50}|$) between original and 50\%-compressed samples, where $l_{Q}$ is the loss at quality $Q$. For clarity, samples are sorted by $l_{100}$. he results clearly demonstrate that samples well-trained by the model (i.e., those exhibiting relatively low loss values) are less sensitive to compression. In contrast, samples that are not learned effectively show significantly higher deviations in loss values compared to their well-trained counterparts.}

\begin{figure}[t!]
    \centering
    \subfigure[Loss]{
    \includegraphics[width=0.45\linewidth]{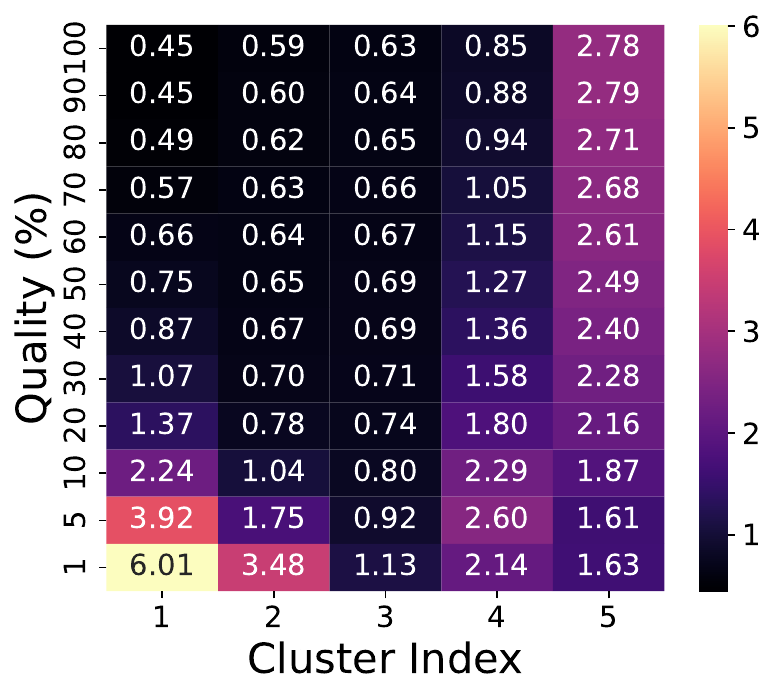}
    }
    \subfigure[$\Delta$Loss]{
    \includegraphics[width=0.45\linewidth]{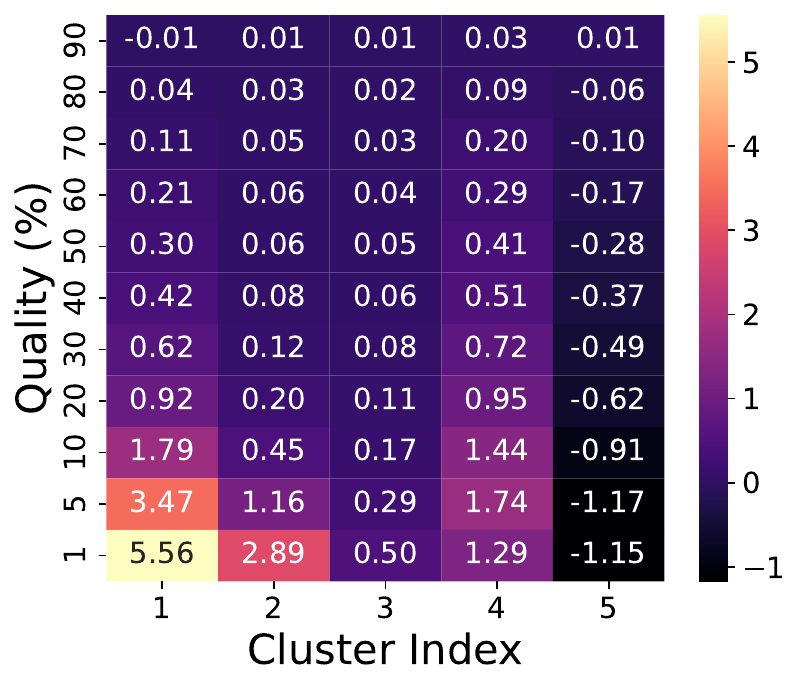}
    }
    \vspace{-3ex}
    \caption{Heatmaps of average loss metrics for test samples grouped into five clusters, evaluated across various compression qualities. (a) The average loss per cluster. (b) The average magnitude of the loss difference per cluster, relative to the original (uncompressed) data.}
    \vspace{-2ex}
    \label{fig:prelim5}
\end{figure}

\rev{Extending these observations further, we computed the loss differences between the original test samples and their compressed counterparts across various compression qualities. We then employed the K-means clustering (with $K=5$ clusters for this experiment) to group these samples based on their compression sensitivity.} Figures~\ref{fig:prelim5} (a) and (b) present heatmaps illustrating the average loss values and average loss differences across different compression qualities, respectively, along with their corresponding cluster assignments.

As Figure~\ref{fig:prelim5} (a) shows, with the exception of Cluster 5, loss values generally tend to increase as the quality of test samples degrades, while each cluster exhibits varying magnitudes and initial degradation thresholds. Specifically, Clusters 1 and 2 demonstrate relatively low loss values for the original high-quality test samples, but exhibit significant loss increases when subjected to the lowest-quality compression. Drawing on observations from Section~\ref{subsec:prelim2}, we infer that these clusters represent samples that are well-learned but predominantly reliant on exclusive features only in high-quality data.

In contrast, Cluster 3 exhibits slightly higher initial loss values compared to Clusters 1 and 2 but lower than those of Clusters 4 and 5, coupled with a slower increase in loss as quality decreases, indicating robustness toward compression. Cluster 4, characterized by the highest average loss among Clusters 1 through 4, shows the steepest increase in loss values, highlighting its heightened sensitivity to compression. Lastly, Cluster 5 demonstrates a unique inverse trend, with decreasing loss values as compression quality deteriorates from initially high levels. This behavior suggests that these samples were initially poorly learned, with the model becoming less confused when extraneous, high-quality features are removed through increased compression.

To summarize, the experiments revealed that \textbf{\textit{each data sample exhibits distinct levels of sensitivity and robustness toward compression}}. Specifically, certain samples minimally influence the model's training effectiveness when compressed, whereas others significantly affect training performance. These findings justify and strongly motivate an adaptive compression strategy that carefully considers the individual importance of samples during the training process.

\subsection{Key Takeaways}
\rev{In summary, our three-part analysis provides a cohesive view of how data compression and storage constraints affect model performance. First, rather than a single fixed balance, the quantity–quality trade-off that yields the best performance can vary with the storage budget. Second, the impact of compression is not uniform across settings; training with intermediately compressed data can yield greater robustness when the test-time quality deviates from the original.  Third, individual samples differ in their sensitivity to compression, with some remaining robust under strong compression while others degrade markedly. Taken together, these results indicate that the optimal balance arises from this underlying differential sensitivity. Thus, a storage-aware system can exploit estimates of sample sensitivity to adapt compression rates per sample and approach the budget-constrained optimum without exhaustive search. We summarize the main findings in Table~\ref{tab:results-summary}.}
\begin{table}[t]
\centering
\footnotesize
\setlength{\tabcolsep}{3pt}
\renewcommand{\arraystretch}{1.0}
\begin{tabularx}{\linewidth}{l X l}
\toprule
\textbf{Item} & \textbf{One-line finding} & \textbf{Evidence} \\
\midrule
RQ1 & Optimal quantity--quality balance \emph{depends on the budget} (no fixed rule). & Fig.~1-2 \\
RQ2 & Compression effect is \emph{non-uniform} & Fig.~3-4 \\
RQ3 & \emph{Per-sample} compression sensitivity differs; clusters show distinct patterns. & Fig.~5-6 \\
Uniform & Uniform drop/fixed-rate compression is \emph{suboptimal}. & Fig.~1-2,4 \\
\bottomrule
\end{tabularx}
\vspace{-2ex}
\caption{\rev{Summary of findings for the core RQs of this work.}}
\vspace{-4ex}
\label{tab:results-summary}
\end{table}

\section{Discussions, \rev{Limitations} and Future Work}
\label{sec:discussion}
\vspace{-1.25ex}

\noindent\textbf{Towards a Practical Adaptive Policy.} \rev{Our results suggest that a \emph{sample-wise} adaptive compression policy under a storage budget would be desirable. To formulate, for each sample $i$, one would choose a retention variable $z_i\!\in\!\{0,1\}$ and a compression quality $q_i$ to maximize held-out validation performance subject to $\sum_i z_i\, b(q_i)\!\le\! B$. Since directly optimizing this joint quantity-quality choice is computationally expensive, a promising direction is to employ lightweight \emph{sensitivity proxies} that approximate the performance impact of compressing a sample (e.g., prediction uncertainty such as entropy/margin, or gradient-/importance-based scores)~\cite{paul2021deep}. In practice, a system could rank candidate (sample, quality) pairs by a benefit-cost ratio derived from such proxies and select them greedily until the budget is met, assigning higher compression to robust samples while preserving fidelity for sensitive ones. Developing calibrated proxies and assessing the effectiveness of this knapsack-style selection across datasets and models remains important future work toward deployable, storage-aware learning.}

\noindent{\textbf{Generalization to Other Data Modalities \rev{and Datasets.}}} 
\rev{A limitation of our study is its focus on a single dataset (CIFAR-10) and a specific lossy codec (JPEG).  While our analysis aims to surface \emph{principles} for storage-aware learning, the quantity-quality balance and per-sample sensitivity can vary across datasets and tasks.  An immediate extension is to validate the observed patterns on diverse benchmarks (e.g., CIFAR-100, ImageNet), driving/robotics datasets with higher scene complexity (e.g., Cityscapes, KITTI, nuScenes), and other modalities/codecs (e.g., audio/video and non-JPEG schemes); thereby assessing robustness to dataset diversity, resolution, and distributional shifts.}


\vspace{0.2ex}
\noindent{\textbf{Synergy with Other On-Device Learning Paradigms.}} Our storage-aware approach is inherently suited for systems with continuous data streams, creating a natural synergy with paradigms like Federated Learning (FL), Continual Learning (CL), and Active Learning (AL)~\cite{wang2022memory, lee2024detrigger}. Specifically, our approach enhances these methods by providing a more sophisticated data retention policy, for example, by enabling efficient knowledge preservation in CL or managing client data under heterogeneous storage in FL. Integrating quality-quantity balancing into these frameworks presents a significant opportunity to develop more robust and efficient on-device intelligence.

\vspace{0.2ex}
\noindent{\textbf{Addressing the Training Overhead.}} While our primary focus is the storage budget, we acknowledge that retaining compressed data, rather than discarding it, may increase computational overhead during training. We posit this is a manageable trade-off, as on-device training often occurs during device idle periods. Moreover, this overhead could be actively mitigated. Future work could explore hybrid policies that combine data compression with selective dropping or employ data consolidation techniques to form more efficient training batches, thereby optimizing the balance between storage efficiency and computational cost~\cite{zhao2020dataset}.

\section{Conclusion}
\label{sec:conclusion}
\vspace{-1.25ex}
This paper's primary contribution is the formalization and empirical characterization of storage-constrained on-device learning. Our analysis reveals that the optimal trade-off between data quantity and quality is dynamic and that individual samples exhibit differential sensitivity to compression. These findings invalidate naive, one-size-fits-all data retention policies and establish the feasibility of an adaptive, sample-wise approach. By defining this problem domain, we lay the essential groundwork for a new class of storage-aware learning systems.
\vspace{-3ex}
\section*{Acknowledgements}
\vspace{-1.25ex}
This work was supported by the Korean MSIT (IITP-2024-2020-0-01461 (ITRC) and IITP-2022-0-00420).


\bibliographystyle{elsarticle-num}
\vspace{-0.3cm}
\bibliography{reference,eis-lab} 
\end{document}